\documentclass[conference]{IEEEtran}
\IEEEoverridecommandlockouts

\usepackage{cite}
\usepackage{amsmath,amssymb,amsfonts}
\usepackage{algorithmic}
\usepackage{graphicx}
\usepackage{textcomp}
\usepackage{xcolor}
\usepackage[font=normalsize]{subfig}
\usepackage{url}
\def\BibTeX{{\rm B\kern-.05em{\sc i\kern-.025em b}\kern-.08em
    T\kern-.1667em\lower.7ex\hbox{E}\kern-.125emX}}
\begin{document}


\title{Cued Speech Generation Leveraging a Pre-trained Audiovisual Text-to-Speech Model \\
\thanks{This work, as part of the Comm4CHILD project, has received funding from the European Union’s Horizon 2020 research and innovation programme under the Marie Sklodowska-Curie Grant Agreement No 860755.}
}


\author{\IEEEauthorblockN{Sanjana Sankar, Martin Lenglet, Gérard Bailly, Denis Beautemps, Thomas Hueber}
\IEEEauthorblockA{\textit{Univ. Grenoble Alpes, CNRS, Grenoble INP, GIPSA-lab, 38000 Grenoble, France} \\
firstname.lastname@grenoble-inp.fr}
}

\maketitle

\begin{abstract}
This paper presents a novel approach for the automatic generation of Cued Speech (ACSG), a visual communication system used by people with hearing impairment to better elicit the spoken language. We explore transfer learning strategies by leveraging a pre-trained audiovisual autoregressive text-to-speech model (AVTacotron2). This model is reprogrammed to infer Cued Speech (CS) hand and lip movements from text input. Experiments are conducted on two publicly available datasets, including one recorded specifically for this study. Performance is assessed using an automatic CS recognition system. With a decoding accuracy at the phonetic level reaching approximately 77\%, the results demonstrate the effectiveness of our approach.
\end{abstract}


\section{Introduction}
\label{sec:intro}

Automatic Cued-Speech Generation (ACSG) involves automatically generating a visual representation of cued speech from text. This visual representation could take the form of either a photorealistic video of a cuer or an animated avatar. ACSG can be seen as the counterpart of Cued-Speech Recognition (ACSR) \cite{sankar22_icassp, sankar23_interspeech}. It is essentially a sequence-to-sequence task, mapping a sequence of discrete units (e.g., text) to a sequence of continuous spatial data (e.g., a video or a sequence of control parameters for a deformable model). Naturally, ACSG shares challenges with Text-to-Speech (TTS) systems and can be considered a specific case of audio-visual TTS (AV-TTS) \cite{AVTacotron2, AVTacotron2_ml}. In this work, we build on recent advancements in AV-TTS, particularly neural encoder-decoder architectures, to address the problem of ACSG.

Despite its similarities to TTS and AV-TTS, ACSG presents unique challenges. First, it is truly a low-resource problem, with only a few available corpora. To the best of our knowledge, for French Cued Speech (LfPC) and continuous CS (as opposed to isolated words), the largest publicly available corpus is CSF22\footnote{\url{https://zenodo.org/records/8392608}}, which contains just a thousand utterances, amounting to roughly one hour of recorded data. Moreover, unlike conventional audiovisual speech data that combines speech recordings with videos of the talker’s face and lips \cite{hsu2010comprehensive}, there is very little CS data ‘in the wild’ on the web that can be easily curated.

Second, unlike AV-TTS, segmenting CS data at the phonetic level (i.e., alignment) is a time-consuming task that is difficult to automate. While phonetic segmentation of lip movements can be somewhat derived from the audio speech signal (excluding anticipatory lip gestures), hand movements in CS follow entirely different dynamics. The onset of hand movements can precede those of the lips by up to 120 ms \cite{attina2004pilot}, and this ‘asynchrony’ between hand and lip movements in CS likely varies depending on the cuer’s proficiency and the communication context.
Interestingly, autoregressive TTS models like Tacotron2 \cite{tacotron2}, and their extensions to AV-TTS \cite{AVTacotron2}, which use encoder-decoder architectures and (cross) attention mechanisms, are designed to learn the mapping between phonetic (or character) sequences and the speech signal without the need for time-alignment. This inspired us to adapt an autoregressive TTS model for ACSG.

Thus, the goals of the present work are threefold: 1) significantly expand the available CS data (for French) by recording a new dataset and making it publicly available, 2) adapt an autoregressive AV-TTS model \cite{AVTacotron2, AVTacotron2_ml} capable of learning to align CS data with text, and 3) explore potential transfer learning strategies from AV-TTS to ACSG to address the low-resource challenge. The dataset, source code, and the trained model will be released along with the publication of this paper.

\begin{figure}[h]
  \centering
  \includegraphics[width=0.8\linewidth]{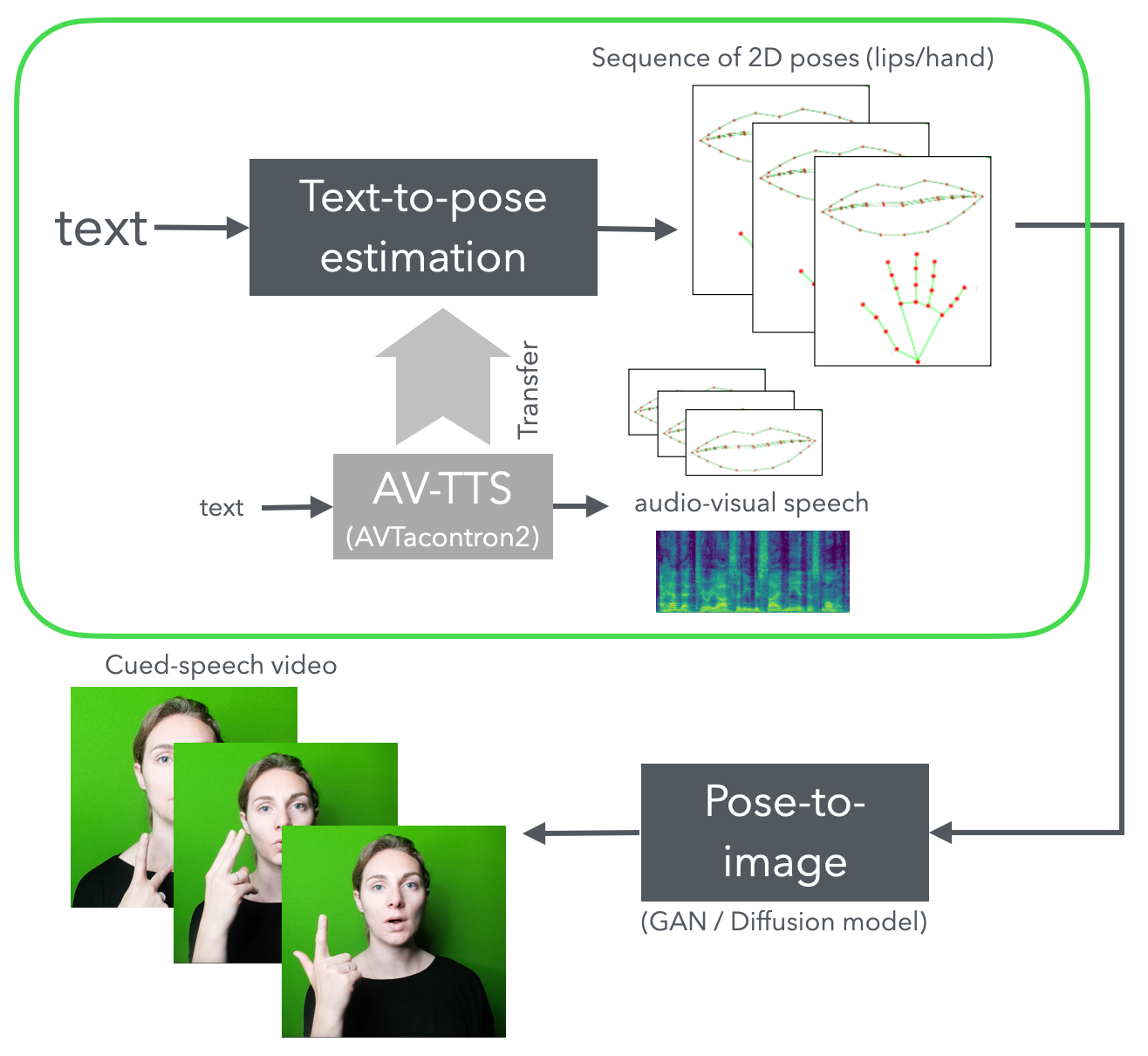}
  \caption{Proposed 2-step framework for the automatic generation of cued-speech from text. The present work focuses on the highlighted green part.}
  \label{fig:gen}
\end{figure}

\section{Related work}
Previous studies have made initial attempts at CS gesture generation. However, the literature on this topic is limited, primarily focusing on multi-modal feature extraction and generation methods. Duchnowski et al. \cite{duchnowski1998} discuss a computerized cueing system using Automatic Speech Recognition (ASR) to determine and display cues to the receiver, with manually selected keywords mapped to pre-defined hand templates. Experiments involved superimposing hand shapes on video recordings of CS tokens, with subjects identifying shapes without audio cues. Gibert et al. \cite{guillaume} focus on 3D movement analysis of the head, face, and hand in CS using multiple cameras and tracking markers, with a method involving "guided-PCA" for disentangling the motion of the lips, jaw and the head. A statistical model of hand articulation was developed, addressing complexities like wrist and finger phalanges' influence on marker positions. Govokina et al. \cite{govokina} explore HMM for gesture synthesis and segmentation in LfPC with improved synchronization of gesture durations with acoustic phoneme durations. Elisei et al. \cite{elisei} present a method for capturing data and creating 3D models to analyze CS and synthesize audiovisual content. 
Being also a gesture-based communication system, ACSG shares some challenges with sign language generation \cite{bangham2000virtual, cox2002tessa, zwitserlood2005synthetic, efthimiou2012dicta,stoll2020text2sign}. Unlike CS, sign language is not built on acoustic speech, as there is no direct one-to-one correspondence between phonemes or words and visual cues, except in the case of certain lexical signs. Finally, to the best of our knowledge, only one recent study (preprint) has investigated the use of deep learning techniques for ACSG in Chinese \cite{lei2024bridge}. Notably, the input text is combined with a textual description of the target hand and lip gestures, and both are used together to generate CS data using a diffusion model. While this approach could have served as a potential baseline, it was designed for (and evaluated on) a different language (Chinese CS), and we leave a comparison with our approach for future work.

\section{Material and Methods}
\label{sec:format}

 In our approach to ACSG, we break down the problem into two main stages, as illustrated in Fig. \ref{fig:gen} (a similar approach is described in \cite{stoll2020text2sign} for sign language generation). The first stage involves translating text or phonetic sequences into a sequence of hand and lips poses, each pose being a set of landmarks describing the position and shape of each articulator. The second stage, which we plan to explore in future work, will focus on generating photorealistic videos or potentially developing an avatar capable of performing cued speech based on these poses. This paper concentrates on the first phase, where we generate 2D poses of hand and lips from text by reprogramming an existing AV-TTS model. Since the output at this stage is not yet visually user-friendly, perceptual evaluation is not feasible. Therefore, we evaluate the system using an existing automatic cued speech recognition (ACSR) system (see our previous work \cite{sankar22_icassp}). Importantly, we investigate different transfer learning strategies for addressing the problem of lack of CS data.


\subsection{Datasets}
\label{sec:datasets}
The previously available datasets for French Cued Speech (CS) include CSF18V2 \cite{csf18v2} and CSF22 \cite{csf22}. The CSF18V2 dataset, though linguistically comprehensive for French Cued Speech (CS), is limited by its short duration and low video resolution (720 x 576 pixels), which hinders effective training for ACSG. High video quality is crucial for our purposes because our regions of interest are the lips and hands. Lips, especially, occupy relatively small areas of the frame. The greater the number of pixels available to represent these features, the better the system's ability to accurately capture and reproduce the fine details necessary for effective ACSG. The CSF22 dataset offers higher resolution but suffers from a variable framerate, making it less suitable for generation tasks. 

To address these limitations, we developed a new dataset, hereafter referred to as CSF23, which ensures a consistent framerate and high image quality. This was achieved through the implementation of a custom recording software, which facilitated the display of prompts for participants to cue and record. The experimental setup includes LED lights (placed next to the cuer’s face) and a green background. The cuer was also requested to utter the sentences while cueing. The sentences for the prompts were taken from discourses at the French National Assembly and were recorded by a professional cuer with typical hearing. The dataset consists of 2,654 videos (one per sentence), encompassing 66,664 phonetic samples and totaling 3.5 hours of data, recorded at a stable frame rate of 30 fps and FullHD resolution (1920x1080). The audio was recorded at 44.1 kHz. To the best of our knowledge, upon release, this will be the most extensive (single cuer) dataset available for continuous French CS. 


\subsection{Model}
\begin{figure*}
  \centering
  \includegraphics[width=1\textwidth]{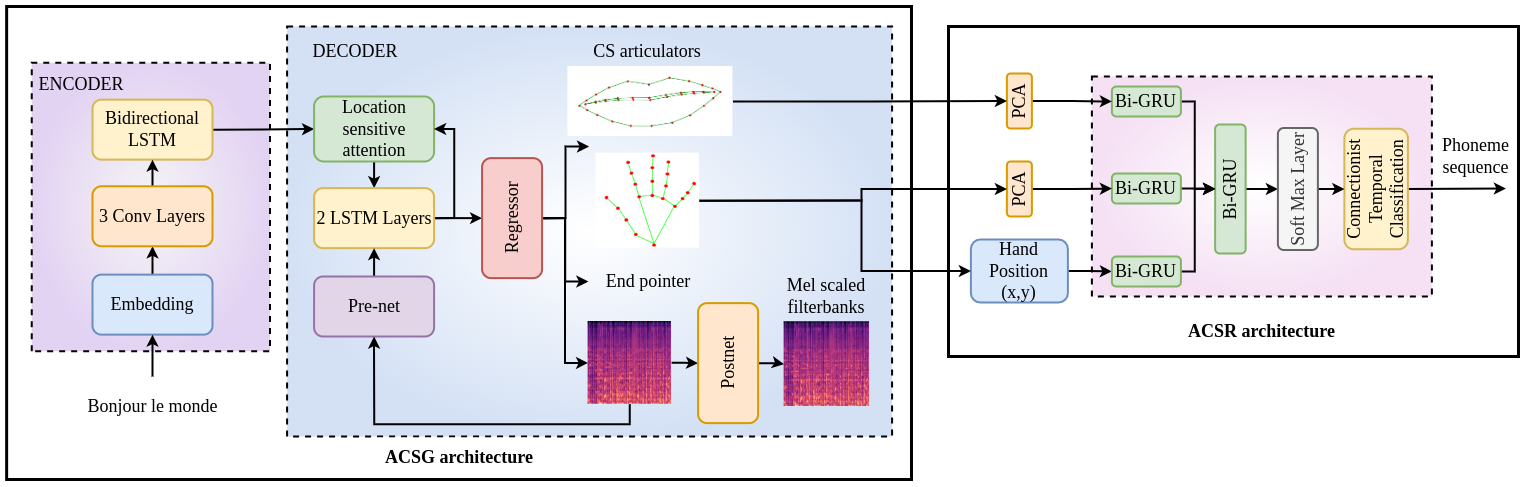} 
  \caption{The ACSG architecture is a modified AVTacotron2 with an additional regression layer to generate both mel-spectrogram and the CS articulators (hand and lips) and the ACSR architecture is the pre-trained model from \cite{sankar22_icassp} used for evaluating the generated features.}
  \label{fig:avtacotron2_cs}
\end{figure*}

The present work aims to adapt an Audiovisual Tacotron2 (AVTacotron2), an AV-TTS model \cite{AVTacotron2}, for the ACSG task. 
The AVTacotron2 is presented in Fig. \ref{fig:avtacotron2_cs}. It is an extension of the original Tacotron2 model, a state-of-the-art RNN-based encoder-decoder model with attention mechanisms widely used for audio speech synthesis. The main difference with the Tacotron2 is the addition of a regressor to predict visual features. This regressor is a linear projection layer following the second LSTM layer of the autoregressive decoder. The visual features are predicted independently of the autoregressive process and does not involve any integration with it. Additionally, no postnet is applied to the visual features, ensuring that they remain distinct from the audio (mel-spectrogram) processing pipeline. This design enables the synchronization of the speech acoustics with visual features, effectively combining these modalities.

\subsection{Transfer learning}

In this study, we investigate the benefit of transferring representation, learned while performing AV-TTS, to ACSG. 
Here, we seek to bootstrap ACSG by first learning the relationship between text as input, and both lips and audio as output, on a dataset dedicated to this task, avoiding dealing with the complex asynchrony problem posed by the combination of hand and lips in CS. 
To that purpose, we used the publicly available AV-TTS model for French recently released by Lenglet et al. \cite{AVTacotron2_ml}.\footnote{\url{https://github.com/MartinLenglet/AVTacotron2.git}}  
This model was trained on a large high-quality audiovisual dataset (the video focusing on the speaker's face), consisting of isolated extracts from French Novels and French parliament debates (6,538 utterances in total), uttered by a single French speaker (different from the one used in the CS datasets considered in this study). More information about the dataset can be found in \cite{AVTacotron2_ml} and implementation details about the model are given in Section \ref{sec:implem}. 


We investigate the 3 following different strategies leveraging this model for ACSG: 

\begin{itemize}
    \item \textbf{S1 (no finetuning)}: First, we train an AVTacotron2 model entirely from scratch on our CS data (text as input, and lips, hand and audio as output) to establish a baseline for performance. Here, the training process optimizes the parameters of the encoder, attention mechanism, and decoder jointly. 
    \item \textbf{S2 (warm-starting from a pretrained model)}: Here, we load the pre-trained AVTacotron2 model described above, while keeping the visual regressor in its initial state. Then, in the second stage, we trained all the model components (encoder, decoder and regressors) on the CS data. 
    \item \textbf{S3 (training with a frozen pre-trained encoder)}: Here, we followed the same procedure as S2 but we fine-tuned the model with CS data, i.e. both visual  (hand and lips) and audio, while keeping the encoder frozen. 
    \end{itemize}

\subsection{Implementation Details}
\label{sec:implem}
For all datasets, the audio speech signal (resampled at 22,050 Hz) was encoded into a 80-band mel spectrogram using a window size of 1,024 samples and a hop size of 256 samples. 
For the visual features, we used the \textit{Mediapipe} toolkit \cite{mediapipe} to extract automatically the sequence of hand and lips poses from each CS video. Each lip (resp. hand) pose was defined as 42 (resp. 21) 2D landmarks. Pose analysis was done at the same rate as the mel-spectrogram extraction, ensuring the same number of audio and visual observations for each utterance. Finally, visual features were reduced using principal component analysis (PCA), retaining the first 10 components, which captured up to 99\% of the variance, for each stream (lips and hand), resulting in a 20-dimensional vector.

As for the pre-trained AV-TTS system \cite{AVTacotron2_ml}, it follows the same architecture as the one reported in \cite{AVTacotron2} (the latter being not publicly available). In short, its encoder is made of 3 convolutional layers (512 filters, kernel size set to 5), 
followed by a bidirectional LSTM layer  (512 cells). 
The representation generated by the encoder are then processed by a location-sensitive attention module which learns to align the input text with the visual output features (in addition to audio). The attention weights are processed by a 1D CNN layer (32 filters, kernel size set to 31). The resulting context vector is used by the decoder. The latter consists of 2 unidirectional LSTM layers (1024 cells each), a prenet (fully connected layer with a dimension of 256), a linear regressor that generates the continuous mel-spectrogram frames and corresponding visual features, and finally a postnet comprises 5 convolutional layers (512 filters, kernel size set to 5), followed by a linear layer (80 units). The model is trained end-to-end using the Adam optimizer with a learning rate of 0.001. The loss function combines MSE loss for the mel-spectrogram prediction, hand-feature prediction, lip-feature prediction, and gate loss to ensure accurate prediction of the end of the sequence. As for the data, we performed a train-test split with 90\% of the data allocated to the training set and 10\% to the test set to evaluate the model’s performance. 

\subsection{Evaluation using ACSR}
\label{sec:acsr}
Since the output of our ACSG system at this stage is a 2D hand and lips poses, and not visually decodable representation (e.g. a photorealistic image or an avatar), it is hard to evaluate the performance of the system using perceptual test with CS human experts. Therefore, we propose to assess the phonetic content of the estimated hand and lip movements using an automatic cued-speech recognition system (ACSR). To do this, we use the system we proposed in our previous study \cite{sankar22_icassp}, the architecture of which is shown in Fig. \ref{fig:avtacotron2_cs}. This system takes in the 2 streams of estimated visual features (hand and lips), and processed them by a set of Bi-directional Gated Recurrent Unit (Bi-GRU). Resulting embeddings are then concatenated and fed to another Bi-GRU, followed by final \textit{softmax} layer. This network is trained using a CTC (Connectionist Temporal Classification) loss function \cite{GravesCTCdecoder}, which allows for the direct decoding of the most likely sequence of phonemes. 
Since the temporal dynamics of the generated features are different from those of the original features\footnote{As demonstrated by Lenglet et al. \cite{thesis_ml}, the phoneme durations in Tacotron2 are encoded by the encoder. Since the encoder is frozen during the fine-tuning process, this temporal dynamic remains consistent with that of the pre-trained model.}, we needed to fine-tune the last layer of the ACSR model with the generated data for it to learn this new temporal dynamics. 
Later, we report the performance of the ACSR decoder when processing generated CS visual data (hand and lips) in terms of 'accuracy', noted $Acc$ and defined as $Acc. (\%)= (N-D-S-I) /\/ N$, where $N$, $D$, $S$, $I$ are the number of phonemes in the test set, deletions, substitutions and insertions. 

\section{Results}
\begin{figure}
    \centering
    \includegraphics[width=1\linewidth]{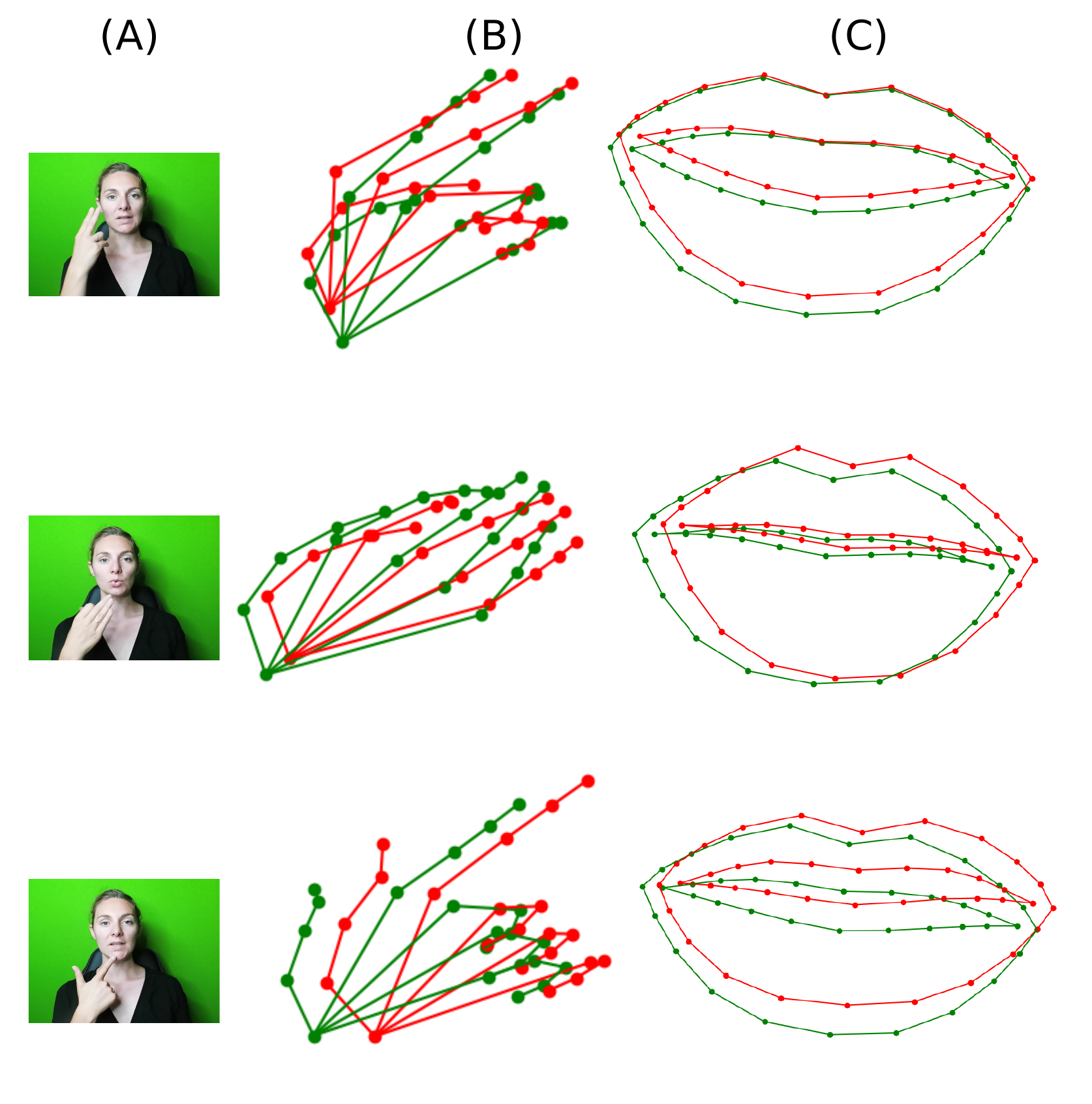}
    \caption{Column (A) shows the original frame from the CS video. Columns (B) and (C) show the generated (in green) and expected (in red) hand and lips features resp.}
    \label{fig:avt2_gen}
\end{figure}

Training the AVTacotron2 from scratch (strategy S1) and warm-starting the model with the weights from the pre-trained audiovisual model (strategy S2)  generated acceptable sequences for the hand, however, they did not yield good results for the lips and often generated an average lip shape for every frame. The only successful strategy was S3, i.e. freezing the encoder of the pre-trained AV-TTS model (lips and audio) and fine-tuning on both visual and audio CS data (hand, lips and audio). In Fig. \ref{fig:avt2_gen}, we show some examples of generated hand and lip movements using this strategy (time correspondence with ground truth data was obtained using dynamic time warping). However, the fact that the S2 strategy failed reveals a catastrophic forgetting problem; the intermediate-level representation learned from the textual input within the encoder, becomes inaccurate when the CS data is taken into account. In the following, we discuss the results obtained following either S1 or S3 strategies (i.e. no fine-tuning, and fine-tuning with pre-training on audio and freezing the encoder). 
As mentioned in Sec. \ref{sec:acsr}, an automatic phonetic decoder of CS (ACSR in Fig. \ref{fig:avtacotron2_cs}) is used to assess the accuracy of the generated hand and lips trajectories at the phonetic level. Without fine-tuning (S1), we obtained an accuracy $Acc.$ of only $17.06\%$ vs. $77.3\%$ with the fine-tuning strategy S3. While there is of course room for improvement (especially since several deletion and insertion errors remain, likely due to the absence of a language model in our ACSR system), the result confirms our working hypothesis that ACSG can benefit from a pretraining on audiovisual speech data (lips and audio).  
Then, since the new recorded dataset CSF23 share some common linguistic material with the audiovisual dataset used to pre-train the AVTacotron2 model, we ran a control experiment on the CSF22 dataset which does not present such overlap. Interestingly, we obtained an accuracy at the phonetic level of $71\%$, i.e. a difference of only $6\%$ w.r.t to CSF23. This shows that the linguistic overlap does introduce a bias, but this bias remains limited and the proposed approach generalizes relatively well to unseen sentences. Finally, from Fig. \ref{fig:avt2_gen} it can be seen that the overall shape of both the hand and lip movements is generally reproduced well, although there is an occasional offset of the features, it does not significantly affect the ability of an expert to decode CS. 

\section{Conclusions and Perspectives}
In conclusion, this paper introduces a novel approach to ACSG by reprogramming an AV-TTS neural model (AVTacotron2) to generate synchronized hand and lip movements from text input. The experimental results demonstrate that this method effectively generates visual features, achieving promising recognition accuracy when evaluated with an ACSR system. Leveraging the encoder block from a pretrained AV-TTS model proves effective in training the model for Cued Speech. Finally, further development is necessary to advance from 2D poses of hand and lips to either avatar-based implementations or the synthesis of photo-realistic videos using, for instance, GAN \cite{goodfellow2014generative} or Diffusion Models \cite{sohl-dickstein2015deep}. Such advancements will be crucial in enabling an effective mode of communication for individuals relying on cued speech.

\vfill
\newpage

\bibliographystyle{IEEEbib}
\bibliography{refs}

\begin{thebibliography}{10}

\bibitem{sankar22_icassp}
Sanjana Sankar, Denis Beautemps, and Thomas Hueber,
\newblock ``Multistream neural architectures for cued speech recognition using a pre-trained visual feature extractor and constrained ctc decoding,''
\newblock in {\em Proc. ICASSP}, 2022, pp. 8477--8481.

\bibitem{sankar23_interspeech}
Sanjana Sankar, Denis Beautemps, Frédéric Elisei, Olivier Perrotin, and Thomas Hueber,
\newblock ``{Investigating the dynamics of hand and lips in French Cued Speech using attention mechanisms and CTC-based decoding},''
\newblock in {\em Proc. of Interspeech 2023}, 2023, pp. 4978--4982.

\bibitem{AVTacotron2}
Ahmed Hussen~Abdelaziz, Anushree~Prasanna Kumar, Chloe Seivwright, Gabriele Fanelli, Justin Binder, Yannis Stylianou, and Sachin Kajareker,
\newblock ``Audiovisual speech synthesis using {T}acotron2,''
\newblock in {\em Proc. ICMI}, 2021, p. 503–511.

\bibitem{AVTacotron2_ml}
Martin Lenglet, Olivier Perrotin, and G{\'e}rard Bailly,
\newblock ``Fastlips: an end-to-end audiovisual text-to-speech system with lip features prediction for virtual avatars,''
\newblock in {\em Proc. of Interspeech}, 2024.

\bibitem{hsu2010comprehensive}
John Hsu and Chia-Hsiu Lee,
\newblock ``A comprehensive review of audiovisual speech recognition and synthesis,''
\newblock {\em Journal of Speech, Language, and Hearing Research}, vol. 53, no. 2, pp. 1111--1124, 2010.

\bibitem{attina2004pilot}
V.~Attina, D.~Beautemps, M.-A. Cathiard, and M.~Odisio,
\newblock ``A pilot study of temporal organization in cued speech production of french syllables: Rules for a cued speech synthesizer,''
\newblock {\em Speech Communication}, vol. 44, no. 1, pp. 197--214, 2004.

\bibitem{tacotron2}
Jonathan Shen, Ruoming Pang, Ron~J Weiss, Mike Schuster, Navdeep Jaitly, Zongheng Yang, Zhifeng Chen, Yu~Zhang, Yuxuan Wang, RJ~Skerry-Ryan, et~al.,
\newblock ``Natural tts synthesis by conditioning wavenet on mel spectrogram predictions,''
\newblock in {\em Proc. of ICASSP}, 2018, pp. 4779--4783.

\bibitem{duchnowski1998}
Paul Duchnowski, Louis~D Braida, Darlene Lum, Marcia Sexton, Julie Krause, and Shashi Banthia,
\newblock ``Automatic generation of cued speech for the deaf: status and outlook,''
\newblock in {\em Proc. of AVSP}, 1998.

\bibitem{guillaume}
Guillaume Gibert, Gérard Bailly, Denis Beautemps, and Frédéric Elisei,
\newblock ``Analysis and synthesis of the 3d movements of the head, face and hand of a speaker using cued speech,''
\newblock {\em Journal of the Acoustical Society of America}, vol. 118, no. 2, pp. 1144--1153, 2005.

\bibitem{govokina}
Oxana Govokhina,
\newblock {\em Modèles de génération de trajectoires pour l’animation de visages parlants},
\newblock Ph.d. dissertation, Institut National Polytechnique de Grenoble - INPG, 2008.

\bibitem{elisei}
Frédéric Elisei, Gérard Bailly, Guillaume Gibert, and Rémi Brun,
\newblock ``Capturing data and realistic 3d models for cued speech analysis and audiovisual synthesis,''
\newblock in {\em Proc. of AVSP}, 2005, pp. 125--130.

\bibitem{bangham2000virtual}
J.~A. Bangham, S.~J. Cox, R.~Elliott, J.~R.~W. Glauert, I.~Marshall, S.~Rankov, and M.~Wells,
\newblock ``Virtual signing: Capture, animation, storage and transmission – an overview of the visicast project,''
\newblock {\em Speech and Language Processing for Disabled and Elderly People}, 2000.

\bibitem{cox2002tessa}
Stephen Cox, Mike Lincoln, Judy Tryggvason, Melanie Nakisa, Mark Wells, Marcus Tutt, and Sanja Abbott,
\newblock ``Tessa, a system to aid communication with deaf people,''
\newblock in {\em Proc. of the 5th international ACM conference on assistive technologies}, 2002, pp. 205--212.

\bibitem{zwitserlood2005synthetic}
Inge Zwitserlood, Margriet Verlinden, Johan Ros, Sanny Van Der~Schoot, and T~Netherlands,
\newblock ``Synthetic signing for the deaf: Esign,''
\newblock in {\em Proc. of CVHI}, 2004, vol.~3.

\bibitem{efthimiou2012dicta}
Eleni Efthimiou, Stavroula-Evita Fotinea, Thomas Hanke, John Glauert, Richard Bowden, Annelies Braffort, Christophe Collet, Petros Maragos, and Fran{\c{c}}ois Lefebvre-Albaret,
\newblock ``The dicta-sign wiki: Enabling web communication for the deaf,''
\newblock in {\em Proc. of International Conference on Computers for Handicapped Persons (ICCHP)}, 2012, pp. 205--212.

\bibitem{stoll2020text2sign}
Stephanie Stoll, Necati~Cihan Camgoz, Simon Hadfield, and Richard Bowden,
\newblock ``Text2sign: Towards sign language production using neural machine translation and generative adversarial networks,''
\newblock {\em International Journal of Computer Vision}, vol. 128, no. 4, pp. 891--908, 2020.

\bibitem{lei2024bridge}
Wentao Lei, Li~Liu, and Jun Wang,
\newblock ``Bridge to non-barrier communication: Gloss-prompted fine-grained cued speech gesture generation with diffusion model,''
\newblock arXiv, (preprint), https://arxiv.org/abs/2404.19277, 2024.

\bibitem{csf18v2}
Li~Liu, Thomas Hueber, Gang Feng, Denis Beautemps, and Sanjana Sankar,
\newblock ``{CSF18},'' [Dataset, https://doi.org/10.5281/zenodo.5554849], may 2022.

\bibitem{csf22}
Sanjana Sankar, Denis Beautemps, and Thomas Hueber,
\newblock ``{CSF22},'' [Dataset, https://doi.org/10.5281/zenodo.8392608], September 2023.

\bibitem{mediapipe}
Camillo Lugaresi, Jiuqiang Tang, Hadon Nash, Chris McClanahan, Esha Uboweja, Michael Hays, Fan Zhang, Chuo{-}Ling Chang, Ming~Guang Yong, Juhyun Lee, Wan{-}Teh Chang, Wei Hua, Manfred Georg, and Matthias Grundmann,
\newblock ``Mediapipe: {A} framework for building perception pipelines,''
\newblock {\em CoRR}, vol. abs/1906.08172, 2019.

\bibitem{GravesCTCdecoder}
Alex Graves,
\newblock {\em Supervised Sequence Labelling with Recurrent Neural Networks},
\newblock Springer Berlin Heidelberg, 2012.

\bibitem{thesis_ml}
Martin Lenglet,
\newblock {\em {Analysis of Latent Representations of Neural Text-To-Speech Models for Expressive Audio-Visual Synthesis}},
\newblock Theses, {Universit{\'e} Grenoble Alpes}, Dec. 2023.

\bibitem{goodfellow2014generative}
Ian Goodfellow, Jean Pouget-Abadie, Mehdi Mirza, Bing Xu, David Warde-Farley, Sherjil Ozair, Aaron Courville, and Yoshua Bengio,
\newblock ``Generative adversarial nets,''
\newblock in {\em Advances in neural information processing systems}, 2014, pp. 2672--2680.

\bibitem{sohl-dickstein2015deep}
Jascha Sohl-Dickstein, Eric Weiss, Niru Maheswaranathan, and Surya Ganguli,
\newblock ``Deep unsupervised learning using nonequilibrium thermodynamics,''
\newblock in {\em Proc. of ICML}, 2015, pp. 2256--2265.

\end{thebibliography}

\end{document}